\documentclass[10pt,twocolumn,letterpaper]{article}

\pdfoutput=1

\usepackage{cvpr}
\usepackage{times}
\usepackage{epsfig}
\usepackage{graphicx}
\usepackage{amsmath}
\usepackage{amssymb}

\usepackage{tabularx}
\usepackage{multirow}
\usepackage{caption} 
\usepackage{subcaption}
\usepackage{arydshln}

\usepackage{array}
\usepackage{xcolor}
\usepackage{booktabs}
\usepackage{siunitx}

\usepackage[pagebackref=true,breaklinks=true,letterpaper=true,colorlinks,bookmarks=false]{hyperref}

\cvprfinalcopy 

\def\cvprPaperID{5775} 

\DeclareMathOperator*{\argmin}{arg\,min}
\DeclareMathAlphabet\mathbfcal{OMS}{cmsy}{b}{n}

\newcolumntype{C}[1]{>{\centering\let\newline\\\arraybackslash\hspace{0pt}}m{#1}}

\begin{document}

\title{Multiview-Consistent Semi-Supervised Learning for 3D Human Pose Estimation}

\author{Rahul Mitra$^{1}$\thanks{- equal contribution}, Nitesh B. Gundavarapu$^{2*}$, Abhishek Sharma$^3$, Arjun Jain$^{3,4}$ \\
$^1$IIT Bombay $^2$University of California, San Diego $^3$Axogyan AI, Bangalore $^4$ IISc Bangalore
}
\maketitle

\begin{abstract}
The best performing methods for 3D human pose estimation from monocular images require large amounts of in-the-wild 2D and controlled 3D pose annotated datasets which are costly and require sophisticated systems to acquire. To reduce this annotation dependency, we propose Multiview-Consistent Semi Supervised Learning (MCSS) framework that utilizes similarity in pose information from unannotated, uncalibrated but synchronized multi-view videos of human motions as additional weak supervision signal to guide 3D human pose regression. Our framework applies hard-negative mining based on temporal relations in multi-view videos to arrive at a multi-view consistent pose embedding. When jointly trained with limited 3D pose annotations, our approach improves the baseline by $25$\% and state-of-the-art by $8.7$\%, whilst using substantially smaller networks. Lastly, but importantly, we demonstrate the advantages of the learned embedding and establish view-invariant pose retrieval benchmarks on two popular, publicly available multi-view human pose datasets, Human 3.6M and MPI-INF-3DHP, to facilitate future research.
\end{abstract}

\section{Introduction}
Over the years, the performance of monocular 3D human pose estimation has improved significantly due to increasingly sophisticated CNN models ~\cite{zhou-weak,pav-vol,drpose3d,integral,compositional,vnect,tome2017lifting}. For training, these methods depend on the availability of large-scale 3D-pose annotated data, which is costly and challenging to obtain, especially under in-the-wild setting for articulated poses. The two most popular 3D-pose annotated datasets, Human3.6M~\cite{h36m_dataset} (3.6M samples) and MPI-INF-3DHP~\cite{mpi-inf-3dhp} (1.3M samples), are biased towards indoor-like environment with uniform background and illumination. Therefore, 3D-pose models trained on these datasets don't generalize well for real-world scenarios~\cite{dabral, zhou-weak}.

Limited training data, or costly annotation, poses serious challenges to not only deep-learning based methods, but other machine-learning methods as well. Semi-supervised approaches~\cite{semisurvey, semi2, semi3, semi4} have been extensively used in the past to leverage large-scale unlabelled datasets along with small labelled dataset to improve performance. Semi-supervised methods try to exploit the structure/invariances in the data to generate additional learning signals for training. Unlike classical machine-learning models that use fixed feature representation, deep-learning models can learn a suitable feature representation from data as part of training process too. This unique ability calls for semi-supervised approaches to encourage better features representation learning from large-scale unlabelled data for generalization. Intuitively it’s more appealing to leverage semi-supervised training signals that are more relevant to the final application. Therefore, given the vast diversity of computer-vision tasks, it remains an exciting area of research to innovate novel semi-supervision signals.
\begin{figure*}[t]
\begin{subfigure}[]{0.75\textwidth}
                \centering
                \includegraphics[width=0.9\linewidth, height=0.45\linewidth]{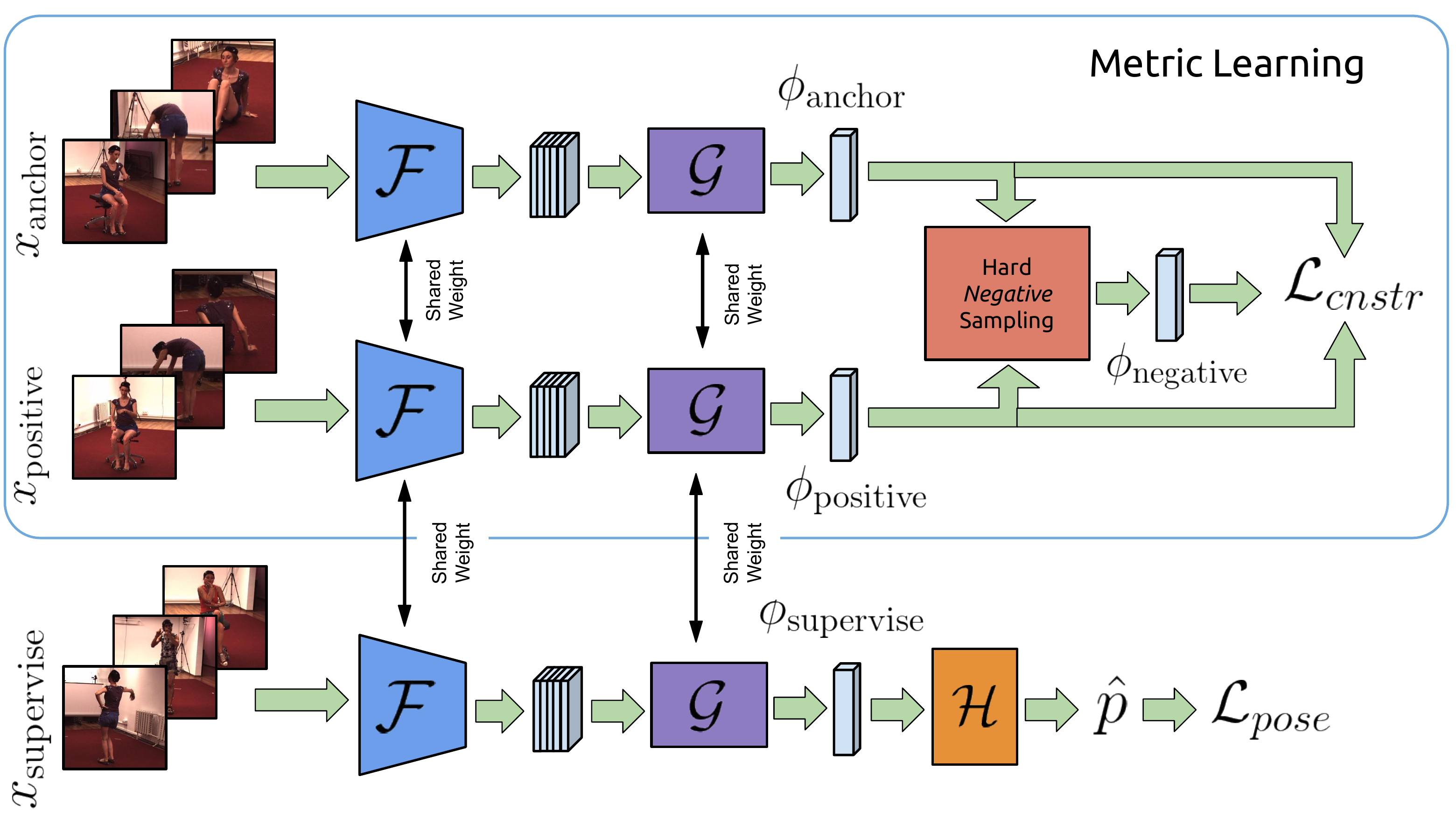}
                \caption{}
                \label{fig:desp-net}
\end{subfigure} \hspace{-1em}
\begin{subfigure}[]{0.25\textwidth}
        \centering
        \includegraphics[width=0.8\linewidth]{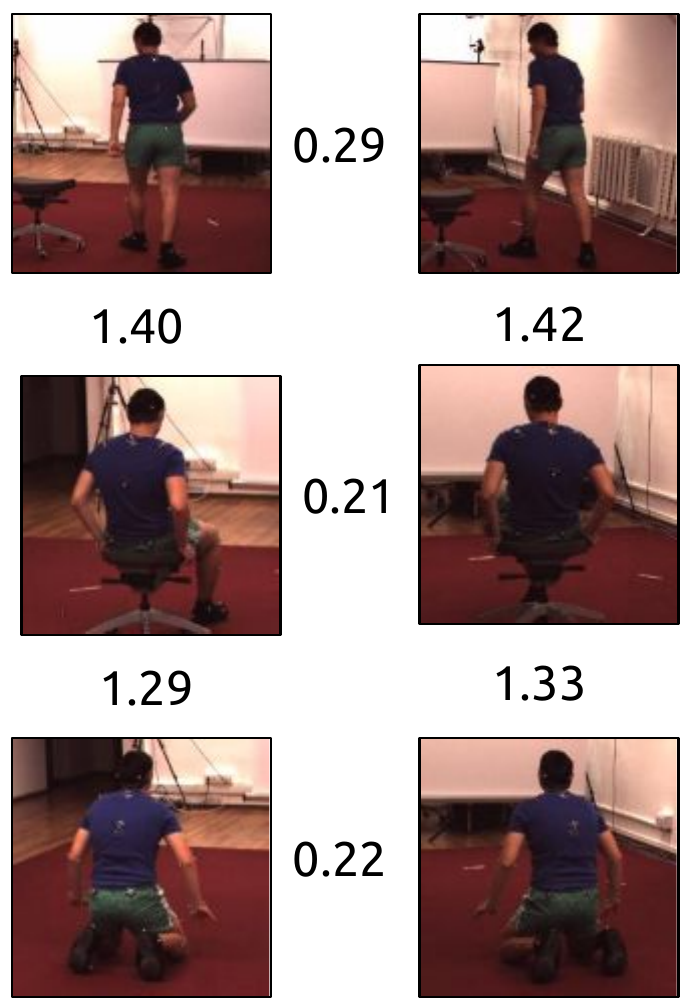}
        \vspace{2.3em}
        \caption{}
        \label{fig:emb-dist-qualitative}
\end{subfigure}
\vspace{-.2cm}
\caption{\textbf{(a)} Training framework for learning our pose embedding and subsequent \emph{canonical} pose estimation. $x_{\text{anchor}}$ and $x_{\text{positive}}$ are a batch of \emph{anchor} and \emph{positive} image pairs taken from different camera views. $x_{\text{supervise}}$ is the batch of images with 3D-pose supervision. $\mathbfcal{F}$ is the ResNet based feature extractor. $\mathbfcal{G}$ maps features extracted from $\mathbfcal{F}$ to our embedding $\phi$. The \emph{Hard Negative Sampling} module performs in-batch hard mining as given in Eq.~\ref{eq:in-batch-neg}. Module $\mathbfcal{H}$ regresses pose $\hat{p}$ from our embedding~$\phi$. \textbf{(b)} Distances between a few images in our learned embedding space. Each column represents images in the same pose from different view. Images across rows have different poses. The numbers between a pair of images represent its embedding distance. The distance is low for pairs with the same pose irrespective of viewpoint and high for those having different poses.}   
\end{figure*}

To this end, we leverage projective multiview consistency to create a novel metric-learning based semi-supervised framework for 3D human-pose estimation. Multiview consistency has served as a fundamental paradigm in computer vision for more than 40 years and gave rise to some of the most used algorithms such as stereo~\cite{stereo}, structure from motion~\cite{sfm}, motion capture~\cite{mocapsurvey}, simultaneous localization and mapping~\cite{slam}, etc. From human-pose estimation perspective, the intrinsic 3D-pose of the human-body remains the same across multiple different views. Therefore, a deep-CNN should \emph{ideally} be able to map 2D-images corresponding to a common 3D-pose, captured from different viewpoints, to nearby points in an embedding space. Intuitively, such a deep-CNN is learning feature representations that are invariant to different views of the human-pose. Therefore, we posit that perhaps it can learn to project 2D images, from different viewpoints, into a \emph{canonical 3D-pose} space in $\mathcal{R}^N$. In Fig.~\ref{fig:emb-dist-qualitative}, we show a few embedding distances between different images from the Human3.6M~\cite{h36m_dataset} and provide empirical evidence to the aforementioned hypothesis via a novel cross-view pose-retrieval experiment. Unfortunately, embedding-vectors, $x$, from such a space do not translate directly to the 3D coordinates of human-pose. Therefore, we learn another transformation function from embedding to pose space  and regress with small 3D-pose supervision while training. Since, the embedding is shared between the pose-supervision and semi-supervised metric-learning, it leads to better generalizeable features for 3D-pose estimation. We name our proposed framework as \emph{Multiview Consistent Semi-Supervised} learning, or \textbf{MCSS} for short.

The proposed framework fits really well with the practical requirements of our problem because it's relatively easier to obtain real-world time-synchronized video streams of humans from multiple viewpoints \vs setting up capture rigs for 3D-annotated data out in-the-wild. An alternative approach could be to setup a calibrated multi-camera capture rig in-the-wild and use triangulation from 2D-pose annotated images to obtain 3D-pose. But, it still requires hand-annotated 2D-poses or an automated 2D-pose generation system . In~\cite{kocabas-cvpr19}, a pre-trained 2D-pose network has been used to generate \emph{pseudo} 3D-pose labels for training a 3D-pose network. Yet another approach exploits relative camera extrinsics for cross-view image generation via a latent embedding~\cite{rhodin-eccv}. We, on the other hand, don't assume such requirements to yield a more practical solution for the limited data challenge.

We use MCSS to improve 3D-pose estimation performance with limited 3D supervision. In Sec.~\ref{sec:pose-eval}, we show the performance variation as 3D supervision is decreased.  Sec.~\ref{sec:view-inv-pose-retrieval} demonstrates the richness of view-invariant MCSS embedding for capturing human-pose structure with the help of a carefully designed cross-view pose-retrieval task on Human3.6M and MPI-INF-3DHP to serve as a benchmark for future research in this direction. A summary of our contributions is,
\begin{itemize}
  \setlength{\itemsep}{1pt}
  \setlength{\parskip}{0pt}
  \setlength{\parsep}{0pt}
    \item Proposed a novel Multiview-Consistent Semi-Supervised learning framework for 3D-human-pose estimation.
    \item Achieved state-of-the-art performance on Human 3.6M dataset with limited 3D supervision.
    \item Formulated a cross-view pose-retrieval benchmark on Human3.6M and MPI-INF-3DHP datasets.
\end{itemize}

\section{Related Work}
This section first reviews prior approaches for learning human-pose embedding followed by a discussion of previous weakly supervised methods for monocular 3D human pose estimation to bring out their differences with our approach.

\subsection{Human Pose Embedding}
Historically, human-pose embeddings have been employed in tracking persons~\cite{emb-tracking-2,emb-tracking}. Estimation of 3D human pose and viewpoint from input silhouettes via learning a low dimension manifold is shown in~\cite{emb-siloh}. 2D-pose regression and retrieval by pose similarity embedding is shown in~\cite{thin-slice, pose-match}, but they require 2D-pose labels. In \cite{pose-emb-video}, the need for 2D-pose labels is eliminated by using human motion videos and temporal ordering as weak supervision in a metric learning framework. Unlike the aforementioned approaches, we learn a 3D-pose embedding by leveraging intrinsic 3D-pose consistency from synchronized multi-view videos. In~\cite{salzman-ijcv}, a 3D-pose embedding is learnt using an over-complete auto-encoder for better structure preservation, but it still requires requires 3D-pose labels for the entire dataset.

\subsection{Weakly Supervised 3D Human Pose Estimation}
Majority of supervised 3D-pose Estimation algorithms~\cite{vnect,martinez,tome2017lifting,deep-human-cvpr,integral,zhao-cvpr19, compositional, zhou-weak, saurabh} require 3D-pose labels in conjunction with either 2D-pose labels or a pre-trained 2D pose estimator to learn a deep-CNN mapping from images to 3D-pose or images to 2D-pose followed by 2D-to-3D lifting task. Some methods refine these pose estimates using either temporal cues, anthropometric constraints, geometric constraints or additional supervision \cite{dabral, selfrep1, selfrep2, selfrep3, selfrep4, selfrep5}. A complete decoupling between 2D and 3D pose estimation is presented in \cite{selfrep2} with the use of generative lifting network followed by a back-projection constraint to achieve generalization. Another line of work focuses on augmenting 2D/3D-pose labels using mesh representation \cite{kanazawa1, kanazawa2,selfrep3, shapeloop} or a dense pose representation \cite{densepose, holopose} to improve pose estimation. All the aforementioned approaches require large amount of annotated 2D and/or 3D labels while our method is designed for limited 3D-pose labels only.

\textbf{Strong 2D and limited/no 3D supervision} 
In recent years, weak-supervision from limited 3D-pose labels along with in-the-wild 2D-pose labels has gained popularity, because labelling 2D-pose  is easier than labelling 3D-pose \cite{videopose3d, kocabas-cvpr19,rhodin-cvpr,rhodin-like-2d,siameq,tyagi-cvpr19}. 
A weak-supervision in the form of re-projection constraints on the predicted 3D pose is proposed ~\cite{videopose3d}. Mostly, such approaches take advantage of multi-view images during training by means of geometric constraints \cite{rhodin-cvpr,kocabas-cvpr19, tyagi-cvpr19}, domain adaptation and adversarial constraints \cite{tyagi-cvpr19}, or cross-view reprojection constraints \cite{rhodin-like-2d}. In~\cite{rhodin-like-2d}, a latent 3D-pose embedding is learned by reconstructing 2D-pose from the embedding in a different view. A shallow network with limited 3D-pose supervision is learned to regress 3D-pose from the embedding. A network with pre-trained weights for 2D-pose estimation is used for 3D-pose estimation in~\cite{rhodin-cvpr} followed by multi-view geometric consistency loss. Pseudo 3D-pose labels are generated in ~\cite{kocabas-cvpr19} for training, while adversarial losses between the 2D skeleton and re-projection of predicted 3D-pose on different views is used for learning in ~\cite{tyagi-cvpr19}. In \cite{siameq}, starting with 2D pose inputs, a  lifting network is trained with siamese loss on the embedding from multiple views to achieve a weak supervision for 3D-pose. Unlike us, \cite{rhodin-cvpr, tyagi-cvpr19, kocabas-cvpr19} require strong 2D-pose estimation systems trained on MPII or COCO datasets while \cite{videopose3d,siameq,rhodin-like-2d} directly work on 2D-pose detections. We, on the other hand, don't need any 2D-pose labels or pre-trained 2D-pose estimation systems.

\textbf{limited/no 2D and limited 3D supervision} - To alleviate the need for a large amount of 2D-pose labels, \cite{rhodin-eccv} learns an unsupervised geometry aware embedding and estimates 3D-pose from embedding with limited 3D supervision. Novel view synthesis using multi-view synchronized videos of human motions is used to learn a geometry-aware embedding. This method however still requires camera extrinsics and background extraction and performs worse than our approach.

Our approach falls in the same category as we don't use any 2D-pose labels. We utilize synchronized videos from multiple views to learn a pose embedding with limited 3D-pose labels such that similar pose samples are mapped close to each other in the embedding space. Unlike \cite{rhodin-eccv}, we don't require camera extrinsics and background extraction. Moreover, we exploit multiview-consistency to directly obtain a canonical pose instead of  performing image-reconstruction, which affords smaller networks, Resnet-18~\cite{resnet} vs. Resnet-50.

\section{Proposed Approach}
\label{sec:proposed-approach}
Our proposed MCSS approach consists of two modules- i) Multiview-consistent metric-learning from time synchronised videos (Sec.~\ref{sec:metric-learning}) and ii) 3D-pose regression with limited 3D supervision (Sec.~\ref{sec:canonical-pose}). Both the modules are jointly trained as shown in Fig.~\ref{fig:desp-net}. Metric-learning acts as semi- supervision signal to reduces the dependency on large-scale 3D-pose labels while pose-regression encourages to learn pose-specific features. 

\subsection{Multiview-Consistent Metric Learning}
\label{sec:metric-learning}
 We utilize \emph{Hardnet} framework~\cite{hardnet} to learn pose embedding. The datasets used for training is divided into images belonging to one of $\mathcal{S} = \{ S_1, S_2, \dots S_n \}$ set of subjects. $\mathcal{P} \subset {{\rm I\!R}}^{16 \times 3}$ is the set of all possible poses and each pose is viewed from $\mathcal{V} = \{ v_1, v_2, \dots v_{q} \}$ viewpoints. For training \emph{hardnet}, each batch consists of paired \emph{anchor}~$(\mathcal{X}^{v_a}_p(S_i) \in \mathcal{X})$ and \emph{positive}~$(\mathcal{X}^{v_b}_p(S_i) \in \mathcal{X})$ images, from subject $S_i$, with the same pose, $p \in \mathcal{P}$, taken from two different viewpoints $v_a$ and $v_b$, here $\mathcal{X} \subset {{\rm I\!R}}^{3 \times 256 \times 256}$ is the set of images.
 
 We pass both the \emph{anchor} and \emph{positive} images through feature extractor ($\mathcal{F}_{\theta_{\mathcal{F}}} : \mathcal{X} \rightarrow \Psi; \: \Psi \subset {{\rm I\!R}}^{512 \times 4 \times 4}$) to generate features \{$\psi^{v_a}_p$, $\psi^{v_b}_p\} \in \Psi$. The feature extractor network is parameterised by $\theta_{\mathcal{F}}$. The features are then finally passed through an embedding generating network ($\mathcal{G}_{\theta_{\mathcal{G}}} : \Psi \rightarrow \Phi; \Phi \subset {{\rm I\!R}}^{dim_{\phi}}$; where $dim_{\phi}$ is dimension of our embedding). Let's assume we feed \emph{anchor} and \emph{positive} images to $\mathcal{F}$ in batches of $m$. Once corresponding  features $\{\phi^{v_{a_1}}_{p_1}, \dots, \phi^{v_{a_m}}_{p_m}\}$ and $\{\phi^{v_{b_1}}_{p_1}, \dots, \phi^{v_{b_m}}_{p_m}\}$ are computed, we create a distance matrix $D$ of size of $m \times m$ with $D(i, j) \:=\: {\lVert \phi^{v_{a_i}}_{p_i} - \phi^{v_{b_j}}_{p_j} \rVert}_2$. \emph{Negative}s $\phi^{v_{j_{min}}}_{p_{j_{min}}}$ and $\phi^{v_{k_{min}}}_{p_{k_{min}}}$ for each of $\phi^{v_{a_i}}_{p_i}$ and $\phi^{v_{b_i}}_{p_i}$ are then sampled from the current batch which lie closest in the embedding space from $\phi^{v_{a_i}}_{p_i}$ and the $\phi^{v_{b_i}}_{p_i}$ respectively. Mathematically, the sampling is formulated in Eq.~\ref{eq:in-batch-neg}. Here, $\beta$ denotes the minimum distance between a hard-mined \emph{negative} and \emph{anchor/positive} in embedding space. The threshold $\beta$ is necessary for stable training and to avoid similar poses as negatives.
 \begin{equation}
 \label{eq:in-batch-neg}
\begin{split}
j_{min} = \argmin_{j \neq i} \delta(D(i, j)) * D(i, j); \\
k_{min} = \argmin_{k \neq i} \delta(D(k, i)) * D(k, i) \\
\delta(x) = 1 \: \text{if} \: x > \beta, \:\: 0 \: \text{otherwise}\\
D^i_{min} = \min (D(i, j_{min}), D(k_{min}, i)) \\
\end{split}
\end{equation}
The average contrastive loss is given in Eq.~\ref{eq:hardnet-siam}, with $\alpha$ being the margin.
\begin{equation}
\begin{split}
	\mathcal{L}_{cnstr} \:=\: \frac{1}{m} \: \sum_{i\:=\:1}^{m} D(i, i) \: + \: \max (0, \alpha \: - D^i_{min})   
\end{split}
\label{eq:hardnet-siam}
\end{equation}

Note that the above learning framework has the following two objectives, namely, a) to bring the \emph{anchors} and \emph{positives} closer and b) to separate out the \emph{negatives} from \emph{anchors} and \emph{positives}. Intuitively, the goal is to learn embedding that captures 3D-pose information while ignoring irrelevant information, such as subject appearance or background. To this end, we propose the following mini-batch selection mechanism to promote the aforementioned goal:

\subsubsection{Mini-batch Selection}
We compose each mini-batch using \emph{anchor} and \emph{positive} pairs from the same subject, and in many cases with overlapping backgrounds, and the \emph{negatives} are also from the same subject since \emph{Hardnet} chooses the hardest negatives from the same mini-batch. The presented mini-batch selection scheme encourages the resulting embedding to capture pose information while discarding subject-appearance and background features when separating the hardest negatives from \emph{anchors} and \emph{positives}. It's due to the inclusion of same personal-appearance and background in both the negatives and anchor/positives, which cannot be used to separate negatives.
We take care to not include temporally close images in a mini-batch by sub-sampling and appropriately choosing $\beta$. Specific hyper-parameter choices are detailed in supplementary material. In Sec.~\ref{sec:view-inv-pose-retrieval}, we show pose retrieval ability of the learned embedding to show that it has indeed successfully captured 3D-pose information. 

\subsection{Pose Regression}
\label{sec:canonical-pose}
\begin{figure}[h]
\centering
\includegraphics[width=0.9\linewidth]{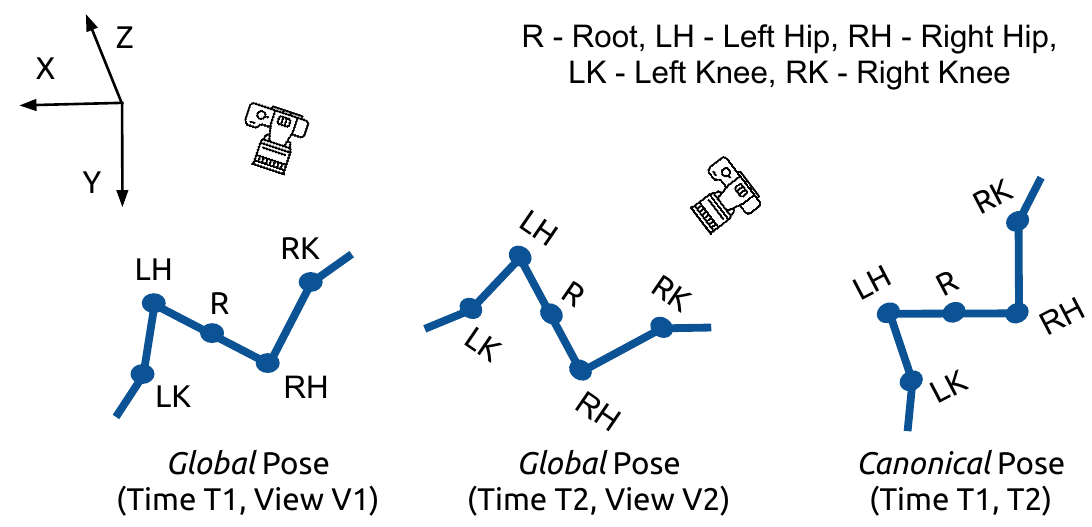}
\caption{Shows top view of bottom half of a human skeleton taken at two different time instants and view points. The left and middle images show two poses having different joint co-ordinates when presented in the \emph{global} pose while having same projections in their respective cameras. On the contrary \emph{canonical} pose provides provides a uniform representation.}  
\label{fig:global-canonical}
\end{figure}

Most 3D-pose estimation approaches focus on regressing for pose in the local camera coordinate system \cite{zhou-weak,pav-vol,tome2017lifting,martinez,integral,selfrep2,rhodin-cvpr}. In our framework, however, 2D-images captured from different views are all mapped to nearby embedding locations, if their intrinsic 3D-poses are the same. Therefore, 3D-pose regression using our embedding is ambiguous because the local camera coordinate system is lost. Moreover, the relation from our embedding to the view-specific 3D-pose is \emph{one-to-many}. In order to address this issue, we make use of the MoCap system's global coordinate to represent the 3D-poses instead of view-specific 3D-poses. Hence, synchronous frames captured from different views are labelled as one global-coordinate 3D-pose. However, asynchronous frames can contain poses which are rigid transformations of one another with same 2D projections. In such cases, the mapping from our embedding to 3D-pose is again an ill-posed \emph{one-to-many} mapping. In Fig.~\ref{fig:global-canonical}, an example of such ambiguity is illustrated.

\subsubsection{Canonical Pose Representation} 
In order to resolve the aforementioned ambiguities, we formulate a multiview-consistent and rigid rotation invariant 3D-pose representation and refer to it as \emph{canonical} pose. 
Canonical pose is obtained by constraining the bone connecting the pelvis to the left hip joint to be always parallel to XZ plane. In Human3.6M dataset, the upward direction is $+$Z axis while XY plane forms the horizontal. Therefore, we rotate the skeleton about the $+$Z axis until the above mentioned bone is parallel to the XZ plane. We don't require any translation since the joint positions are relative to the pelvis. Mathematically, the transformation from \emph{global} to 
\emph{canonical} is given in Eq.~\ref{eq:glb_can},
\begin{align}
    \begin{split}
        \hat{u} \: = \: \frac{p^{glb}_{lh} - p^{glb}_{root}}{\lVert p^{glb}_{lh} - p^{glb}_{root} \rVert}; \;\;
        \hat{u_{xy}} \: = \: \frac{\hat{u_x} \hat{i} + \hat{u_y} \hat{j}}{\lVert \hat{u_x} \hat{i} + \hat{u_y} \hat{j} \rVert}
        \\
         \theta \: = \: \cos^{-1}(\hat{u_{xy}} \cdot \hat{i}) ; \;\;\;
        p^{can} \: = \: R^z_{\theta} * p^{glb} \;\;\;\;\;
    \end{split}
\label{eq:glb_can}
\end{align}
where, $p^{glb}_{root}$ and $p^{glb}_{lh}$ are the root and left-hip joint respectively in the \emph{global} representation. The unit vector along $(p^{glb}_{lh}-p^{glb}_{root})$ is represented as $\hat{u}$, and $\theta$ is the required angle of rotation along the +Z-axis to obtain the canonical pose representation. 
A positive side-effect of canonical pose representation \vs view-specific representation is that our predicted canonical pose doesn't change orientation with variations in camera view. A similar approach to achieve a rotation-invariant pose is suggested in ~\cite{tome2017lifting}. Note that the canonical pose is constructed directly from MoCap system's coordinates and doesn't require camera extrinsics.



Finally, we regress for canonical pose from the latent embedding $\Phi$ with the help of a shallow network ($\mathcal{H}_{\theta_{\mathcal{H}}} : \Phi \rightarrow \mathcal{P}$), as shown in Fig.~\ref{fig:desp-net}. The loss-function is L1-norm between the predicted, $\hat{p}$, and target, $p \in \mathcal{P}$, canonical 3D-pose: $\mathcal{L}_{pose} \:=\: {\lVert p - \hat{p} \rVert}_1$.


\section{Implementation and Training Details}
We use the first 4 residual blocks of an ImageNet~\cite{imagenet} pre-trained ResNet-18 as our backbone. In addition, we modify the batch-norm~(BN) layers by turning off the affine parameters as suggested in \cite{hardnet}. For an input image of size $224 \times 224$ pixels, the output of ResNet is a $512 \times 7 \times 7$ blob, which is further down-sample by 2 using a max-pool operation to get $\Psi$. The embedding network $\mathcal{G}$ is FC layers followed by L2-normalization and it maps $\Psi$ to an embedding of dimension $dim_{\phi}$ (128 in our case), following usual~\cite{hardnet,l2-Net}.

For 3D-pose regression, the input data is normalized for each joint. The pose regression network $\mathcal{G}$ consists of FC layers FC(128, 48), with $\Phi \subset {{\rm I\!R}}^{128}$.  
The margin $\alpha$ for $\mathcal{L}_{conrst}$ is set at $0.6$ and $\beta$ at $0.3$. Adam~\cite{adam} optimized is used with default parameters $(0.9, 0.99)$ with initial learning rate $10^{-3}$. The model is trained for 40 epochs with a drop in learning-rate by 0.1 at every 20 epochs. In our joint training frame work, ratio of the batch size for metric learning to pose regression is kept at $3:1$ with batch size for regression is 22. A schematic diagram of our network architecture is shown in Fig.~\ref{fig:desp-net}.

\subsection{Datasets}
We use the popular Human3.6M ~\cite{h36m_dataset} and MPI-INF-3DHP~\cite{mpi-inf-3dhp} datasets for our experiments.
\begin{itemize}
\setlength{\itemsep}{1pt}
  \setlength{\parskip}{0pt}
  \setlength{\parsep}{0pt}
    \item \textbf{Human3.6M~\cite{h36m_dataset}} contains 3.6 million frames captured from an indoor 
    MoCap system with 4 cameras~$(\mathcal{V})$. It comprises of 11 subjects~$(\mathcal{S})$, each performing 16 actions with each action having 2 sub-actions. Following the standard \emph{Protocol 2} ~\cite{integral}, we use subjects (S1, S5, S6, S7, S8) for training and (S9, S11) for testing. Like several other methods, we also use cropped subjects' using  bounding-boxes provided with the dataset and \textbf{temporal sub-sampling} is done to include every $5^{th}$ and $64^{th}$ frame for training and testing phase, respectively.
    \item \textbf{MPI-INF-3DHP~\cite{mpi-inf-3dhp}} is generated from a MoCap system with 12 synchronized cameras in both indoor and outdoor settings. It contains 8 subjects($\mathcal{S}$) with diverse clothing. We use the 5 chest height cameras($\mathcal{V}$) for both training and test purposes. Since the test set doesn't contain annotated multi-view data, we use S1-S6 for training and S7-S8 for evaluation.
\end{itemize}

\section{Quantitative Evaluation for Pose Estimation}
\label{sec:pose-eval}
We perform the same quantitative experiment as presented in ~\cite{rhodin-eccv} to assess the benefits of the learned embedding in 3D-pose estimation on Human 3.6M dataset. We evaluate using three well adopted metrics, MPJPE, PA-MPJPE and Normalized MPJPE (N-MPJPE) (introduced in ~\cite{rhodin-cvpr}) which incorporates a scale normalization to make the evaluation independent of person's height. We compare our proposed approach and its variants against a baseline which only uses $\mathcal{L}_{pose}$. In addition, we compare our method against the approach proposed by Rhodin et al.~\cite{rhodin-eccv} and \cite{rhodin-cvpr}, although it estimates human poses in the camera coordinate system. We also report the performance of Rhodin et al.~\cite{rhodin-eccv} using ResNet-18 as the feature extractor instead of ResNet-50. It is to be noted that \cite{rhodin-eccv} uses additional information at training time in the form of relative camera rotation and background extraction which requires sophisticated, well calibrated setup. We acknowledge the existence of more accurate methods like ~\cite{tyagi-cvpr19, kocabas-cvpr19, fua-2d} than \cite{rhodin-eccv, rhodin-cvpr} on Human3.6M when abundant 2D and limited 3D labels are available. For comparison with these approaches, however, we report results from ~\cite{rhodin-like-2d} that requires limited 3D supervision but complete 2D supervision from both Human3.6M and MPII~\cite{mpii} dataset. Since, our focus is advancing the research in monocular 3D-pose estimation without using 2D labels under limited 3D-pose labels, we restrict our comparison to cases with limited supervision from both 2D and 3D labels. We don't include the results of~\cite{videopose3d} as it requires multiple temporally adjacent frames at inference-stage and uses pre-trained 2D-pose estimation models learned from large-scale 2D-pose annotated datasets.

\begin{figure}[t]
\begin{center}
\includegraphics[width=0.95\linewidth]{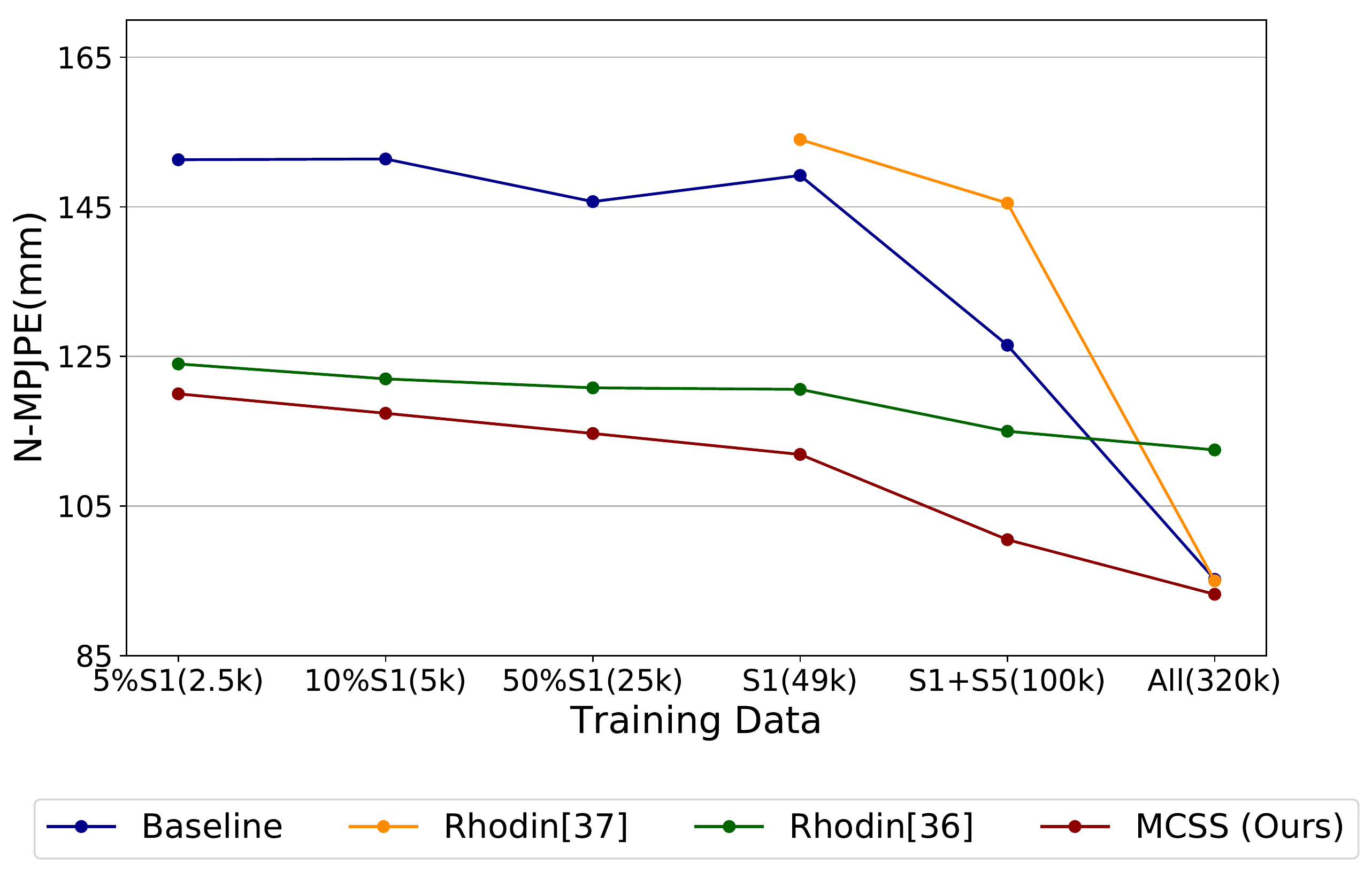}
\end{center}
\vspace{-.5cm}
\caption{
N-MPJPE \vs 3D-pose supervision on test split of Human3.6M. Our proposed model outperforms the baseline and the current state-of-the-art Rhodin et al.~\cite{rhodin-eccv}.}
\label{fig:compare-canonical}
\end{figure}

In order to show performance variation as a function of 3D-pose supervision, we report N-MPJPE values for models trained using different amount of 3D-pose labels, in Fig.~\ref{fig:compare-canonical}. In this experiment, 3D-pose supervision is reduced gradually using all 5 subjects, to S1$+$S5, only S1, 50\% S1, 10\%S1 and finally 5\% S1. MVSS clearly outperforms the baseline by a margin of {\textbf{37.34 N-MPJPE}} when only S1 is used for supervision. Moreover, MVSS degrades gracefully as 3D-pose supervision is reduced, which validates the importance of $\mathcal{L}_{conrst}$ in providing weak supervision to capture 3D-pose. Qualitative comparison of our method against the baseline is shown in Fig.~\ref{fig:compare-canonical}.

In Tab.~\ref{table:can-pose-compare}, we compare MPJPE, N-MPJPE and PA-MPJPE values of our approach against baseline and~\cite{rhodin-eccv}. Clearly, our method outperforms ~\cite{rhodin-eccv} by \textbf{22.4} N-MPJPE when fully supervised on 3D data and by \textbf{10.7} N-MPJPE with 3D-pose supervision limited to S1. For MPJPE however, the margin is \textbf{10.75}. Interestingly as mentioned in \cite{rhodin-eccv}, the performance of \cite{rhodin-cvpr} drastically falls when pre-trained model from strong 2D-pose supervision is not used (reported in Tab.~\ref{table:can-pose-compare} as Rhodin~\cite{rhodin-cvpr}\textsuperscript{*} and Rhodin~\cite{rhodin-cvpr}).

\begin{table}[h]
\centering
\begin{tabular}{ >{\raggedright\arraybackslash}C{10mm}   >{\raggedright\arraybackslash}p{17mm}  >{\raggedleft\arraybackslash}p{10mm}  >{\raggedleft\arraybackslash}p{10mm}  >{\raggedleft\arraybackslash}p{10mm} }
\toprule
 \multicolumn{1}{>{\centering\arraybackslash}C{10mm}}{Supervision} 
    & \multicolumn{1}{>{\centering\arraybackslash}C{17mm}}{Method} 
    & \multicolumn{1}{>{\centering\arraybackslash}C{10mm}}{N-MPJPE}
    & \multicolumn{1}{>{\centering\arraybackslash}C{10mm}}{MPJPE}
    & \multicolumn{1}{>{\centering\arraybackslash}C{10mm}}{PA-MPJPE}\\\midrule
 \multirow{5}{*}{All} 
 & Rhodin~\cite{rhodin-cvpr}\textbf{\textsuperscript{*}} & 63.30 & 66.80 & 51.60 \\
 & Chen~\cite{rhodin-like-2d}\textbf{\textsuperscript{*}} & NA & 80.20 & 58.20 \\ 
 & Baseline & 95.07 & 97.90 & 77.18\\
                       & Rhodin~\cite{rhodin-cvpr} & 95.40 & NA & NA\\
                       & Rhodin~\cite{rhodin-eccv} & 115.00 & NA & NA\\ 
                       & MCSS(Ours) & 92.60 &  94.25 & 72.48 \\

                       \midrule
 \multirow{6}{*}{S1} 
 & Rhodin~\cite{rhodin-cvpr}\textbf{\textsuperscript{*}} & 78.20 & NA & NA \\
 & Chen~\cite{rhodin-like-2d}\textbf{\textsuperscript{*}} & NA & 91.90 & 68.00 \\ 
 & Baseline & 149.28 & 154.78 & 113.69\\
                       & Rhodin~\cite{rhodin-cvpr} & NA & 153.30 & 128.60 \\
                       & Rhodin~\cite{rhodin-eccv} & 122.60 & 131.70 & 98.20\\
                       & Rhodin~\cite{rhodin-eccv}-Res18 & 136.00 & NA & NA \\
                       & MCSS(Ours) & \textbf{111.94} & \textbf{120.95} & \textbf{90.76}\\ 
                       
                       \bottomrule                       
\end{tabular}
\caption{Comparing N-MPJPE and MPJPE values between different approaches on Human 3.6M dataset when supervised on all 5 subjects and on only S1. \textbf{Note:} Pre-trained ImageNet weights are used to initialize the networks by all the methods. Methods or its variants marked with \textbf{`*'} are supervised with large amount of in-the-wild 2D annotations from MPII~\cite{mpii} dataset either during training or by means of a pre-trained 2D pose estimator.  
All other methods use much weaker supervision by assuming no 2D annotations and \emph{MCSS} outperforms the state-of-the-art~\cite{rhodin-eccv} in such settings. NA is assigned against a method if the corresponding result is not reported by the authors.} 
\label{table:can-pose-compare}
\end{table}

\begin{figure*}
\begin{center}
\includegraphics[width=0.98\linewidth]{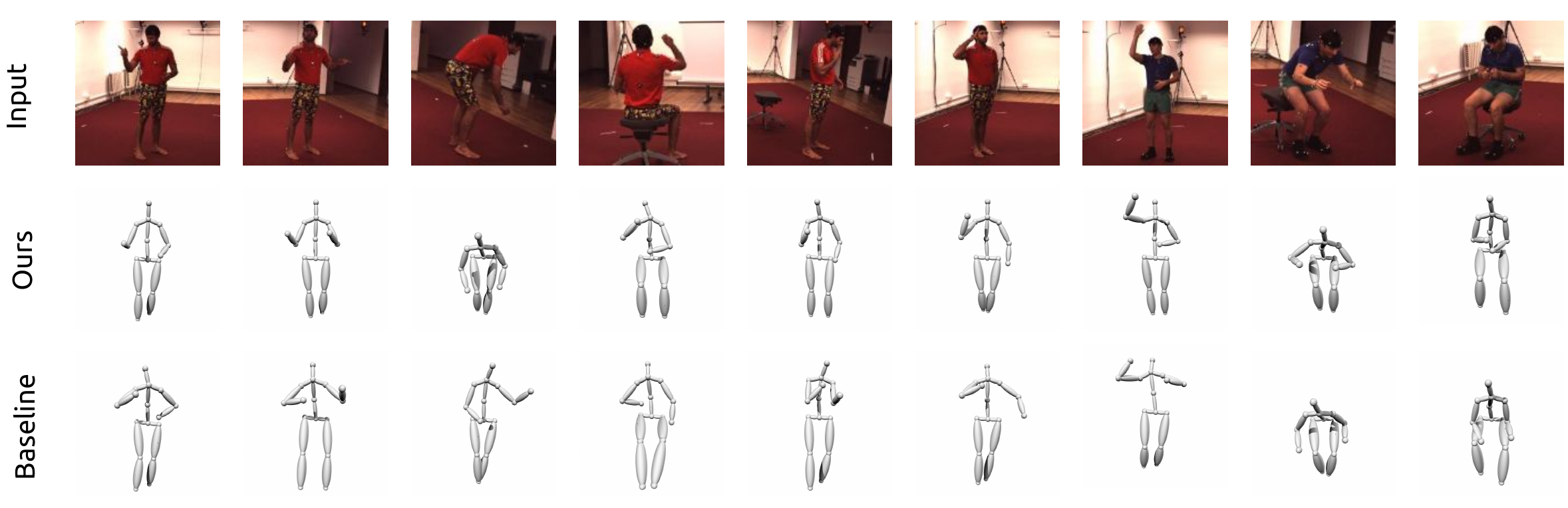}
\vspace{-.5cm}
\end{center}
   \caption{Qualitative results on \emph{canonical} pose estimation by our proposed framework (\textbf{MCSS}) against our \textbf{Baseline} on Human 3.6M test split (S9, S11). Both the models are trained with supervision from labels of subject S1. Our method produces more accurate estimates for even for challenging poses like `sitting' and `bending'.} 
\label{fig:pose-regression-qual}
\end{figure*}

\begin{table}[b]
\centering
\begin{tabular}{>{\raggedright\arraybackslash}p{18mm} >{\raggedright\arraybackslash}p{33mm}  >{\raggedleft\arraybackslash}p{15mm} }
\toprule
 \multicolumn{1}{>{\centering\arraybackslash}C{18mm}}{Supervision} 
    & \multicolumn{1}{>{\centering\arraybackslash}C{33mm}}{Method} 
    & \multicolumn{1}{>{\centering\arraybackslash}C{15mm}}{N-MPJPE}\\ 
    \midrule
 \multirow{2}{*}{S1}   & MCSS & 111.94 \\
                       & MCSS-global & 157.30 \\ 
                       & MCSS-ResNet34 & 115.85 \\ 
    \bottomrule
\end{tabular}
\caption{Comparing N-MPJPE values when pose estimation is done in Mocap's (MCSS-global) and \emph{canonical}(MCSS) representations when only subject S1 is used for supervision. Performance of using ResNet-34 as back-end is reported against MCSS-ResNet34.}
\label{table:can-global-nmpjpe}
\end{table}

As part of ablation studies, we also compare the performance of our learning framework when target pose is represented in MoCap's(\emph{global} pose) against our \emph{canonical} representation in Tab.~\ref{table:can-global-nmpjpe}. We observe dramatic decrease in performance, 45 MPJPE, which validates the importance of canonical representation. We also show the results for a deeper ResNet-34~\cite{resnet} back-end network. We observe a slight drop in performance, 3 MPJPE points, perhaps due to over-fitting.

An additional benefit of our proposed framework is in the use of a much smaller ResNet-18 feature extractor as compared to ResNet-50 used in Rhodin et al~\cite{rhodin-eccv}. This affords an inference time of $24.8$ms \vs $75.3$ms by \cite{rhodin-eccv} on a NVIDIA 1080Ti GPU. Note that Rhodin et al.~\cite{rhodin-eccv} shows degradation in performance when using the smaller ResNet-18 backbone. We attribute it to direct latent embedding similarity learning instead of generative modelling that requires more representation capacity.

\section{Analysis of Learned Embedding}
\label{sec:view-inv-pose-retrieval}
\begin{figure}
    \centering
    \includegraphics[width=.95\linewidth]{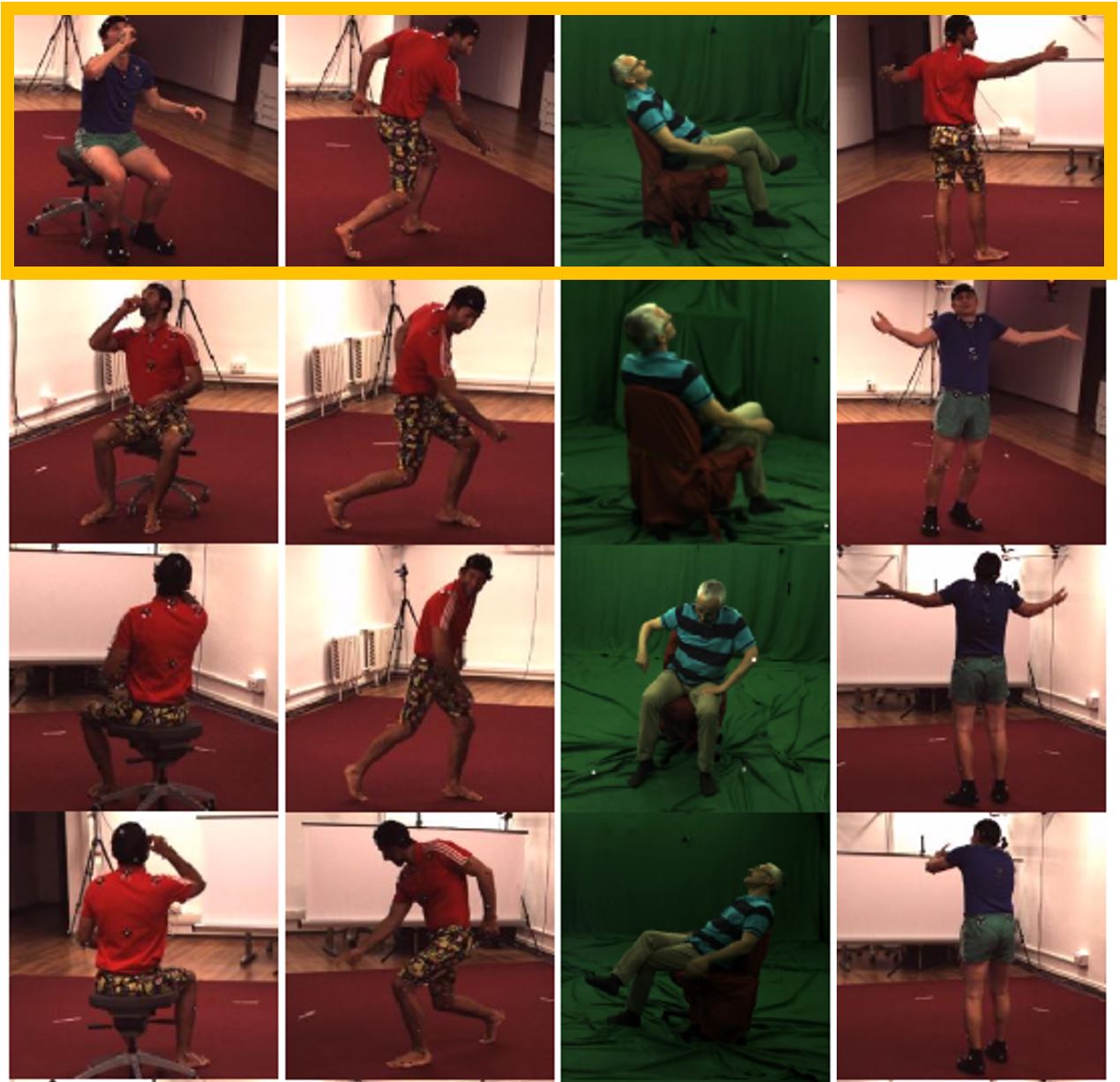}
    \caption{Qualitative image retrievals on Human 3.6M (S9, S11) and MPI-INF-3DHP (S7, S8) test sets. The first row represents query image and the rows below are the top 3 closest images in embedding space. For the left-most and right-most columns, the retrieval database is composed of images from different subject and viewpoint from that of query's. For the middle two columns, retrieval database is composed of images of same subject but different viewpoint from that of query's. Note how the retrieved poses are very similar to query poses.}
    \label{fig:retrieval}
\end{figure}

In this section, we demonstrate the quality of our learned embedding in capturing 3D human-pose by showing i) pose based cluster formation in our embedding space through retrieval tasks, ii) the correlation between embedding and pose distances. In Fig.~\ref{fig:retrieval},  we show qualitative image retrieval results based on embedding distance. We can clearly see that the closest images from other subjects and other viewpoints to the query image in embedding space share similar poses. We additionally provide T-SNE~\cite{tsne} plots of our learned embedding space and experiments on generalization to novel view-point in the supplementary material. 

\subsection{Cross-View and Cross-Subject Pose Retrieval}
Our learned embedding tries to project similar pose-samples close to each other irrespective of the subject, viewpoint and background. To validate this claim, we seek motivation from \cite{pose-emb-video}, \cite{thin-slice} and propose Mean-PA-MPJPE@$K$ to measure the Procrustes Aligned Mean Per Joint Position Error (PA-MPJPE) of $K$ closest neighbours from different views. Since, similar poses in terms of the intrinsic human-body pose can still have different orientations, we use Procrustes Aligned MPJPE to remove this artifact. We compare our model against an \emph{Oracle}, which uses ground truth 3D-pose labels. Given a query image, we ensure that the retrieval database contains images taken from viewpoints other than that of the query image. It is done to clearly bring out the view invariance property of the proposed embedding. 
\begin{figure}[b]
\begin{center}
\includegraphics[width=1.0\linewidth, height=0.7\linewidth]{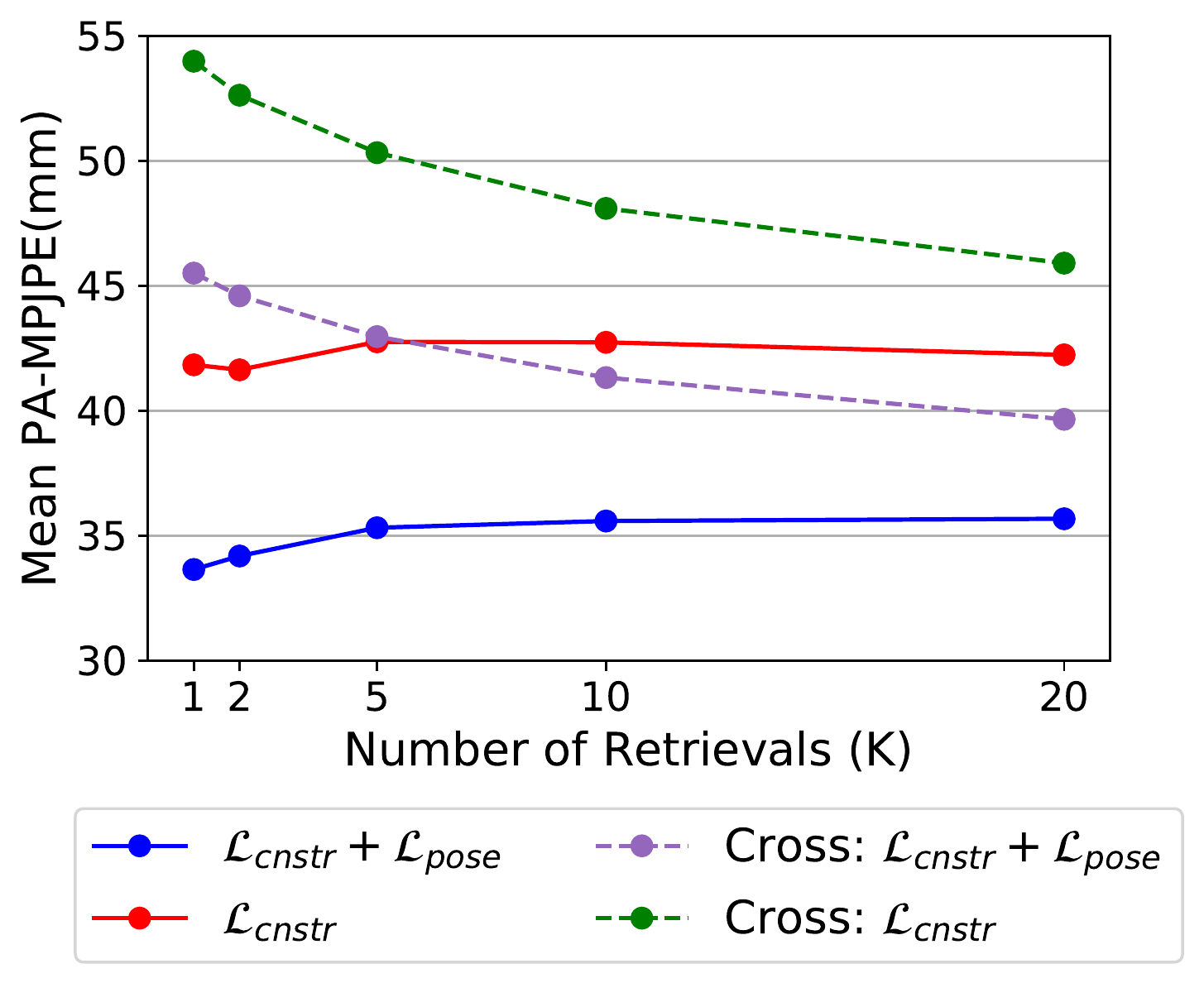}
\end{center}
\caption{Mean-PA-MPJPE for increasing number of retrievals $K$ on Human3.6M dataset. Prefix `Cross-' indicates retrieval done on different subjects from that of query. $\mathcal{L}_{pose}$ is from subject S1 in both the cases. All values reported are relative to an Oracle. Low values indicates our retrieved poses are similar to that of the Oracle. PAMPJPE is in mm.}
\label{fig:mean-pampjpe}
\end{figure}
First, we report the Mean-PA-MPJPE@$K$ between query pose and its $K$ nearest neighbors in the embedding space. In Fig.~\ref{fig:mean-pampjpe}, we show the comparison of Mean-PA-MPJPE@$K$ of retrieved poses when retrieval is done from images with: \\
\textbf{Case 1:} all test subjects including that of query's.\\ 
\textbf{Case 2:} all test subjects except that of query's - \emph{cross}.\\
We report our results relative to the Oracle. The nearly horizontal plots with low errors suggest that our model picks poses similar to that of the Oracle irrespective of $K$. The error rate is slightly higher for $K=1,2$ since our model retrieves images from clusters and does not always pick the one with the lowest error as done by the oracle. The error is lower for \textbf{Case 1} than \textbf{Case 2} due to the presence of images in the query database that share the exact same pose as that of query, but from different viewpoints. We can also note that upon $\mathcal{L}_{pose}$ from S1, the clustering and mean mpjpe improves in both same subject, \textbf{Case 1}, and cross-subject, \textbf{Case 2}, settings falling in line with our expectation that small amount of pose supervision improves clustering.

\begin{table}[!ht]
\centering
\begin{tabular}{>{\raggedright\arraybackslash}p{19mm}  >{\raggedleft\arraybackslash}p{10mm}  >{\raggedleft\arraybackslash}p{10mm} >{\raggedleft\arraybackslash}p{10mm} >{\raggedleft\arraybackslash}p{10mm} }
\toprule
 \multicolumn{1}{>{\centering\arraybackslash}C{19mm}}{Method} 
    & \multicolumn{1}{>{\centering\arraybackslash}C{10mm}}{K=1} 
    & \multicolumn{1}{>{\centering\arraybackslash}C{10mm}}{K=5}
    & \multicolumn{1}{>{\centering\arraybackslash}C{10mm}}{K=10}
    & \multicolumn{1}{>{\centering\arraybackslash}C{10mm}}{K=20}\\ \midrule
 $\mathcal{L}_{cnstr}$ & 48.40  & 62.46 & 56.29 & 55.63 \\
 Cross-$\mathcal{L}_{cnstr}$ & 82.29 & 83.53 & 80.65 &76.00 \\ \bottomrule
 \end{tabular}
 \caption{Mean-PA-MPJPE (mm) for increasing number of retrievals (K) on MPI-INF-3DHP dataset after finetuning with $\mathcal{L}_{cnstr}$. Prefix Cross- indicates retrieval is done on subject other than query's. All values are reported with respect to the Oracle.}
\label{table:mpi-inf-ret}
\end{table}

\subsection{Correlation between Embedding and Pose}
\begin{figure}
    \centering
    \includegraphics[width=\linewidth]{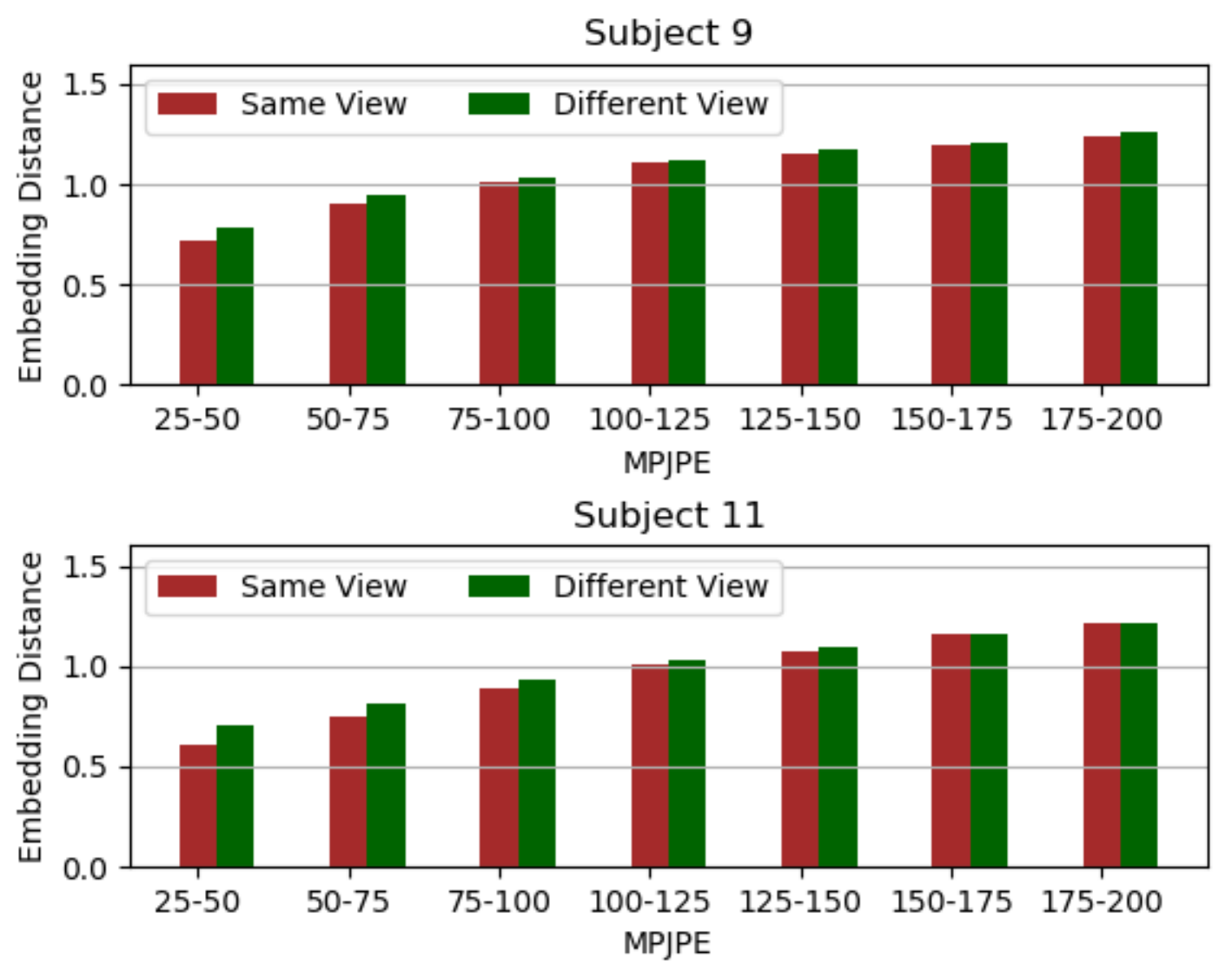}
    \caption{Variation of mean embedding distance with increasing pose variation. We use the show results on (S9, S11) with model being trained with $\mathcal{L}_{cnstr}$ on (S1, S5, S6, S7, S8) and $\mathcal{L}_{pose}$ on (S1). Images are stacked in bins based on the MPJPE difference of their corresponding poses with that of the query. On the Y-axis, the mean embedding distance between the query and the image stacks is plotted. In \textbf{Same View}, the query image and image stacks belong to the same viewpoint while in \textbf{Different View}, the query stacks belong to different viewpoints. The results are averaged over 200 random queries for each subject.}
    \label{fig:emb_dist_pose}
\end{figure}

In this section, we illustrate the variation exhibited by our learned embedding with change in human pose. To this end, we plot mean embedding distance between a query image and stacks of images with increasing pose difference with that of the query in Fig.~\ref{fig:emb_dist_pose}. Both the query and the image stacks belong to the same subject. One can observe a clear positive co-relation between embedding distance and corresponding pose difference. Further, same view and different view show similar correlations with poses justifying the fact that our learned embedding is multi-view consistent.
\subsection{Generalization \& Limitations:}
To test cross-dataset generalization, we applied a model trained on Human 3.6M dataset and performed cross-view pose retrievals on MPI-INF-3DHP dataset. We obtained a mean MPJPE of $119.6$mm and $101.9$mm for $K=10$ and $K=20$ respectively. Further fine-tuning with $\mathcal{L}_{conrst}$ using multi-view images from MPI-INF-3DHP  improved the performance to $62.46$mm and $56.29$mm, see Tab.~\ref{table:mpi-inf-ret}. The dip in performance on cross dataset can be attributed to the fact that our feature extractor and embedding generating network has learnt a mapping from Human 3.6M images to a pose space and the same mapping is not applicable to the domain of MPI-INF-3DHP images because of huge variation in appearance and more challenging variations of poses. However, upon adding $\mathcal{L}_{conrst}$, as shown in  Tab.~\ref{table:mpi-inf-ret} the weak supervision generalizes to new dataset.  


\section{Conclusion and Future Work}
\label{sec:future}
In this paper, we demonstrated a novel Multiview-Consistent Semi-Supervised learning approach to capture 3D human structure for pose estimation and retrieval tasks. With the help of our semi-supervised framework, the need for 3D-pose is reduced. It enables our method to outperform contemporary weakly-supervised approaches even while using a smaller network. Furthermore, we provided strong benchmarks for view-invariant pose retrieval on publicly available datasets.

In future, we plan to use multi-view synchronised videos captured in-the-wild from a larger set of viewpoints to improve generalisation further.  We also plan to extend our framework to capture very fine grained pose variations with our embedding by learning distributions of pose variations in temporally consecutive frames using limited 3D annotations.

{\small

\begin{thebibliography}{10}\itemsep=-1pt

\bibitem{densepose}
R{\i}za Alp~G{\"u}ler, Natalia Neverova, and Iasonas Kokkinos.
\newblock Densepose: Dense human pose estimation in the wild.
\newblock In {\em CVPR}, 2018.

\bibitem{mpii}
Mykhaylo Andriluka, Leonid Pishchulin, Peter Gehler, and Bernt Schiele.
\newblock 2d human pose estimation: New benchmark and state of the art
  analysis.
\newblock In {\em CVPR}, 2014.

\bibitem{selfrep3}
Anurag Arnab, Carl Doersch, and Andrew Zisserman.
\newblock Exploiting temporal context for 3d human pose estimation in the wild.
\newblock In {\em CVPR}, 2019.

\bibitem{slam}
Raja Chatila and Jean-Paul Laumond.
\newblock Position referencing and consistent world modeling for mobile robots.
\newblock In {\em ICRA}, 1985.

\bibitem{tyagi-cvpr19}
Ching-Hang Chen, Ambrish Tyagi, Amit Agrawal, Dylan Drover, Rohith MV, Stefan
  Stojanov, and James~M Rehg.
\newblock Unsupervised 3d pose estimation with geometric self-supervision.
\newblock In {\em CVPR}, 2019.

\bibitem{rhodin-like-2d}
Xipeng Chen, Kwan-Yee Lin, Wentao Liu, Chen Qian, and Liang Lin.
\newblock Weakly-supervised discovery of geometry-aware representation for 3d
  human pose estimation.
\newblock In {\em CVPR}, 2019.

\bibitem{fua-2d}
Xipeng Chen, Kwan-Yee Lin, Wentao Liu, Chen Qian, and Liang Lin.
\newblock Weakly-supervised discovery of geometry-aware representation for 3d
  human pose estimation.
\newblock In {\em CVPR}, 2019.

\bibitem{dabral}
Rishabh Dabral, Anurag Mundhada, Uday Kusupati, Safeer Afaque, Abhishek Sharma,
  and Arjun Jain.
\newblock Learning 3d human pose from structure and motion.
\newblock In {\em ECCV}, 2018.

\bibitem{emb-siloh}
Ahmed Elgammal and Chan-Su Lee.
\newblock Inferring 3d body pose from silhouettes using activity manifold
  learning.
\newblock In {\em CVPR}, 2004.

\bibitem{semisurvey}
Rob Fergus, Yair Weiss, and Antonio Torralba.
\newblock Semi-supervised learning in gigantic image collections.
\newblock In {\em NIPS}, 2009.

\bibitem{holopose}
Riza~Alp Guler and Iasonas Kokkinos.
\newblock Holopose: Holistic 3d human reconstruction in-the-wild.
\newblock In {\em CVPR}, 2019.

\bibitem{selfrep5}
Ikhsanul Habibie, Weipeng Xu, Dushyant Mehta, Gerard Pons-Moll, and Christian
  Theobalt.
\newblock In the wild human pose estimation using explicit 2d features and
  intermediate 3d representations.
\newblock In {\em CVPR}, 2019.

\bibitem{resnet}
Kaiming He, Xiangyu Zhang, Shaoqing Ren, and Jian Sun.
\newblock Deep residual learning for image recognition.
\newblock In {\em CVPR}, 2016.

\bibitem{h36m_dataset}
Catalin Ionescu, Dragos Papava, Vlad Olaru, and Cristian Sminchisescu.
\newblock Human3. 6m: Large scale datasets and predictive methods for 3d human
  sensing in natural environments.
\newblock In {\em TPAMI}, 2013.

\bibitem{kanazawa1}
Angjoo Kanazawa, Michael~J Black, David~W Jacobs, and Jitendra Malik.
\newblock End-to-end recovery of human shape and pose.
\newblock In {\em CVPR}, 2018.

\bibitem{kanazawa2}
Angjoo Kanazawa, Jason~Y Zhang, Panna Felsen, and Jitendra Malik.
\newblock Learning 3d human dynamics from video.
\newblock In {\em CVPR}, 2019.

\bibitem{adam}
Diederik~P Kingma and Jimmy Ba.
\newblock Adam: A method for stochastic optimization.
\newblock In {\em arXiv}, 2014.

\bibitem{kocabas-cvpr19}
Muhammed Kocabas, Salih Karagoz, and Emre Akbas.
\newblock Self-supervised learning of 3d human pose using multi-view geometry.
\newblock In {\em CVPR}, 2019.

\bibitem{sfm}
Jan~J Koenderink and Andrea~J Van~Doorn.
\newblock Affine structure from motion.
\newblock In {\em JOSA A}, 1991.

\bibitem{shapeloop}
Nikos Kolotouros, Georgios Pavlakos, Michael~J Black, and Kostas Daniilidis.
\newblock Learning to reconstruct 3d human pose and shape via model-fitting in
  the loop.
\newblock In {\em CVPR}, 2019.

\bibitem{semi2}
Yevhen Kuznietsov, Jorg Stuckler, and Bastian Leibe.
\newblock Semi-supervised deep learning for monocular depth map prediction.
\newblock In {\em CVPR}, 2017.

\bibitem{thin-slice}
Suha Kwak, Minsu Cho, and Ivan Laptev.
\newblock Thin-slicing for pose: Learning to understand pose without explicit
  pose estimation.
\newblock In {\em CVPR}, 2016.

\bibitem{emb-tracking}
Chan-Su Lee and Ahmed Elgammal.
\newblock Modeling view and posture manifolds for tracking.
\newblock In {\em CVPR}, 2007.

\bibitem{semi4}
Christian Leistner, Helmut Grabner, and Horst Bischof.
\newblock Semi-supervised boosting using visual similarity learning.
\newblock In {\em CVPR}, 2008.

\bibitem{selfrep4}
Chen Li and Gim~Hee Lee.
\newblock Generating multiple hypotheses for 3d human pose estimation with
  mixture density network.
\newblock In {\em CVPR}, 2019.

\bibitem{tsne}
Laurens van~der Maaten and Geoffrey Hinton.
\newblock Visualizing data using t-sne.
\newblock In {\em JMLR}, 2008.

\bibitem{martinez}
Julieta Martinez, Rayat Hossain, Javier Romero, and James~J Little.
\newblock A simple yet effective baseline for 3d human pose estimation.
\newblock In {\em CVPR}, 2017.

\bibitem{mpi-inf-3dhp}
Dushyant Mehta, Helge Rhodin, Dan Casas, Pascal Fua, Oleksandr Sotnychenko,
  Weipeng Xu, and Christian Theobalt.
\newblock Monocular 3d human pose estimation in the wild using improved cnn
  supervision.
\newblock In {\em 3DV}, 2017.

\bibitem{vnect}
Dushyant Mehta, Srinath Sridhar, Oleksandr Sotnychenko, Helge Rhodin, Mohammad
  Shafiei, Hans-Peter Seidel, Weipeng Xu, Dan Casas, and Christian Theobalt.
\newblock Vnect: Real-time 3d human pose estimation with a single rgb camera.
\newblock In {\em TOG}, 2017.

\bibitem{hardnet}
Anastasiia Mishchuk, Dmytro Mishkin, Filip Radenovic, and Jiri Matas.
\newblock Working hard to know your neighbor's margins: Local descriptor
  learning loss.
\newblock In {\em NIPS}, 2017.

\bibitem{mocapsurvey}
Thomas~B Moeslund and Erik Granum.
\newblock A survey of computer vision-based human motion capture.
\newblock In {\em CVIU}, 2001.

\bibitem{pose-match}
Greg Mori, Caroline Pantofaru, Nisarg Kothari, Thomas Leung, George Toderici,
  Alexander Toshev, and Weilong Yang.
\newblock Pose embeddings: A deep architecture for learning to match human
  poses.
\newblock In {\em arXiv}, 2015.

\bibitem{pav-vol}
Georgios Pavlakos, Xiaowei Zhou, Konstantinos~G Derpanis, and Kostas
  Daniilidis.
\newblock Coarse-to-fine volumetric prediction for single-image 3d human pose.
\newblock In {\em CVPR}, 2017.

\bibitem{videopose3d}
Dario Pavllo, Christoph Feichtenhofer, David Grangier, and Michael Auli.
\newblock 3d human pose estimation in video with temporal convolutions and
  semi-supervised training.
\newblock In {\em CVPR}, 2019.

\bibitem{deep-human-cvpr}
Alin-Ionut Popa, Mihai Zanfir, and Cristian Sminchisescu.
\newblock Deep multitask architecture for integrated 2d and 3d human sensing.
\newblock In {\em CVPR}, 2017.

\bibitem{rhodin-eccv}
Helge Rhodin, Mathieu Salzmann, and Pascal Fua.
\newblock Unsupervised geometry-aware representation for 3d human pose
  estimation.
\newblock In {\em ECCV}, 2018.

\bibitem{rhodin-cvpr}
Helge Rhodin, J{\"o}rg Sp{\"o}rri, Isinsu Katircioglu, Victor Constantin,
  Fr{\'e}d{\'e}ric Meyer, Erich M{\"u}ller, Mathieu Salzmann, and Pascal Fua.
\newblock Learning monocular 3d human pose estimation from multi-view images.
\newblock In {\em CVPR}, 2018.

\bibitem{imagenet}
Olga Russakovsky, Jia Deng, Hao Su, Jonathan Krause, Sanjeev Satheesh, Sean Ma,
  Zhiheng Huang, Andrej Karpathy, Aditya Khosla, Michael Bernstein, et~al.
\newblock Imagenet large scale visual recognition challenge.
\newblock In {\em IJCV}, 2015.

\bibitem{stereo}
Steven~M Seitz, Brian Curless, James Diebel, Daniel Scharstein, and Richard
  Szeliski.
\newblock A comparison and evaluation of multi-view stereo reconstruction
  algorithms.
\newblock In {\em CVPR}, 2006.

\bibitem{saurabh}
Saurabh Sharma, Pavan~Teja Varigonda, Prashast Bindal, Abhishek Sharma, and
  Arjun Jain.
\newblock Monocular 3d human pose estimation by generation and ordinal ranking.
\newblock In {\em ICCV}, 2019.

\bibitem{semi3}
Rowland~R Sillito and Robert~B Fisher.
\newblock Semi-supervised learning for anomalous trajectory detection.
\newblock In {\em BMVC}, 2008.

\bibitem{pose-emb-video}
Omer Sumer, Tobias Dencker, and Bjorn Ommer.
\newblock Self-supervised learning of pose embeddings from spatiotemporal
  relations in videos.
\newblock In {\em ICCV}, 2017.

\bibitem{compositional}
Xiao Sun, Jiaxiang Shang, Shuang Liang, and Yichen Wei.
\newblock Compositional human pose regression.
\newblock In {\em ICCV}, 2017.

\bibitem{integral}
Xiao Sun, Bin Xiao, Fangyin Wei, Shuang Liang, and Yichen Wei.
\newblock Integral human pose regression.
\newblock In {\em ECCV}, 2018.

\bibitem{salzman-ijcv}
Bugra Tekin, Isinsu Katircioglu, Mathieu Salzmann, Vincent Lepetit, and Pascal
  Fua.
\newblock Structured prediction of 3d human pose with deep neural networks.
\newblock In {\em BMVC}, 2016.

\bibitem{l2-Net}
Yurun Tian, Bin Fan, and Fuchao Wu.
\newblock L2-net: Deep learning of discriminative patch descriptor in euclidean
  space.
\newblock In {\em CVPR}, 2017.

\bibitem{tome2017lifting}
Denis Tome, Chris Russell, and Lourdes Agapito.
\newblock Lifting from the deep: Convolutional 3d pose estimation from a single
  image.
\newblock In {\em CVPR}, 2017.

\bibitem{emb-tracking-2}
Raquel Urtasun, David~J Fleet, and Pascal Fua.
\newblock 3d people tracking with gaussian process dynamical models.
\newblock In {\em CVPR}, 2006.

\bibitem{siameq}
M{\'a}rton V{\'e}ges, Viktor Varga, and Andr{\'a}s L{\H{o}}rincz.
\newblock 3d human pose estimation with siamese equivariant embedding.
\newblock In {\em Neurocomputing}, 2019.

\bibitem{selfrep2}
Bastian Wandt and Bodo Rosenhahn.
\newblock Repnet: Weakly supervised training of an adversarial reprojection
  network for 3d human pose estimation.
\newblock In {\em CVPR}, 2019.

\bibitem{selfrep1}
Keze Wang, Liang Lin, Chenhan Jiang, Chen Qian, and Pengxu Wei.
\newblock 3d human pose machines with self-supervised learning.
\newblock In {\em TPAMI}, 2019.

\bibitem{drpose3d}
Min Wang, Xipeng Chen, Wentao Liu, Chen Qian, Liang Lin, and Lizhuang Ma.
\newblock Drpose3d: Depth ranking in 3d human pose estimation.
\newblock In {\em IJCAI}, 2018.

\bibitem{zhao-cvpr19}
Long Zhao, Xi Peng, Yu Tian, Mubbasir Kapadia, and Dimitris~N Metaxas.
\newblock Semantic graph convolutional networks for 3d human pose regression.
\newblock In {\em CVPR}, 2019.

\bibitem{zhou-weak}
Xingyi Zhou, Qixing Huang, Xiao Sun, Xiangyang Xue, and Yichen Wei.
\newblock Weakly-supervised transfer for 3d human pose estimation in the wild.
\newblock In {\em ICCV}, 2017.

\end{thebibliography}

}

\end{document}